\documentclass[11pt]{article}

\usepackage{acl}
\usepackage{times}
\usepackage{latexsym}
\usepackage{alltt}
\usepackage[T1]{fontenc}
\usepackage[utf8]{inputenc}
\usepackage{microtype}
\usepackage{float}
\usepackage{xcolor}
\usepackage{hyperref}
\usepackage{tikz}
\usetikzlibrary{shapes, arrows, positioning, calc}
\usepackage{placeins}
\usepackage{fvextra}
\usepackage{amsmath}
\title{BITS Pilani at SemEval-2026 Task 9: Structured Supervised Fine-Tuning with DPO Refinement for Polarization Detection}
\author{
Atharva Gupta$^{1}$ \quad
Dhruv Kumar$^{1}$ \quad
Yash Sinha$^{1}$ \\
$^{1}$Birla Institute of Technology and Science, Pilani, India \\ \newline \\
\texttt{\{f20240519, dhruv.kumar, yash.sinha\}@pilani.bits-pilani.ac.in}
}

\begin{document}
\maketitle
\renewcommand{\thefootnote}{\fnsymbol{footnote}}
\footnotetext[1]{Code available at \url{https://github.com/atharva7-g/POLAR-SemEval-Submission}}
\renewcommand{\thefootnote}{\arabic{footnote}}

\begin{abstract}
The POLAR SemEval-2026 Shared Task aims to detect online polarization and focuses on the classification and identification of multilingual, multicultural, and multi-event polarization. 

Accurate computational detection of online polarization is challenging due to nuanced rhetoric, implicit framing, and the high cost of human-in-the-loop annotation. Building on recent findings that contextual prompting enables large language models to function as strong polarization detectors, we present a two-stage approach for detecting polarization in social media text that combines structured supervised fine-tuning with Direct Preference Optimization (DPO) refinement. 

We fine-tune Qwen 2.5-7B-Instruct with LoRA using an interpretable slot-filling template (target, claim type, manifestation checklist, and justification). We then apply DPO with automatically generated preference pairs to reduce costly false negatives. Our submitted system achieves 0.7664 Macro-F1 on the English test set. Post-submission experiments with Mistral-Nemo-Instruct-2407 and LLM-judge-filtered preference pairs further improve to 0.8162 Macro-F1 (not submitted to CodaBench), surpassing the organiser baseline of 0.7802.
\end{abstract}

\section{Introduction}
\label{sec:intro}

Large Language Models (LLMs) are rapidly being integrated across 
domains, from scientific research to customer service and policy 
analysis \cite{bommasani2022opportunitiesrisksfoundationmodels}. 
Their capacity for large-scale language understanding has made them 
central to tasks such as semantic evaluation, stance detection, and 
content moderation \cite{devlin2019bertpretrainingdeepbidirectional, 
zhang-etal-2024-sentiment, pangtey2025llmsstancedetectionsurvey}.

In particular, LLMs have transformed online discourse analysis by 
enabling fine-grained interpretation of meaning, context, and intent 
at scale, surpassing earlier rule-based and shallow machine learning 
approaches \cite{franceschelli2025creativity, li2024pre}.
Online discourse increasingly reflects strong ideological divides, 
making polarization detection an important task for moderating digital 
communication \cite{Loru_2025}.

In this paper, we study polarization detection in online discourse 
and describe a two-stage approach that combines structured supervised 
fine-tuning with preference-based refinement. Our goal is to improve 
Subtask~1 classification performance, particularly Macro-F1, while 
keeping the system efficient. Our submission targets Subtask~1 
(Polarization Detection) in the English language only; all 
experiments, development evaluations, and the final task submission 
are conducted exclusively on the English portion of the POLAR dataset.

In summary, our contributions are:
\begin{itemize}
    \item We formulate SemEval-2026 Task~9 (POLAR) Subtask~1~\cite{naseem-etal-2026-polar,naseem2026polarbenchmarkmultilingualmulticultural} as 
    binary classification and augment predictions with a structured 
    rationale (target, claim type, manifestation checklist, and 
    justification) using a rigid slot-filling schema to reduce output 
    variance.

    \item We apply Direct Preference Optimization (DPO) 
    \cite{rafailov2024directpreferenceoptimizationlanguage} with 
    automatically constructed preference pairs to discourage overly 
    conservative predictions and reduce false negatives.

    \item We show through ablation that structured reasoning 
    generation and preference-based refinement are mutually 
    reinforcing: the rationale enables more effective DPO pair 
    construction, and post-submission experiments confirm this 
    synergy scales with model capacity.

    \item Post-submission experiments with 
    Mistral-Nemo-Instruct-2407 and LLM-judge-filtered preference 
    pairs improve Macro-F1 to 0.8162, surpassing the organiser baseline of 0.7802.
\end{itemize}

\section{Related Work}
\label{sec:related}

Polarization and related abuse detection are widely studied research 
areas \cite{naseem2026polarbenchmarkmultilingualmulticultural}. 
Earlier shared-task work includes SemEval-2019 HatEval, which 
provides multilingual resources and baselines for related hate and 
abuse detection \cite{basile-etal-2019-semeval}.

Methodologically, recent work explores LLM adaptation via prompting 
and parameter-efficient fine-tuning, finding that fine-tuning often 
outperforms in-context learning on polarization-adjacent tasks 
\cite{maggini2025llmshyperpartisanfakepolarized}. Complementary to 
standard cross-entropy fine-tuning, supervised contrastive objectives 
have been shown to improve robustness and generalization of 
pre-trained language model classifiers, especially in low-data 
settings \cite{gunel2021scl}. \citet{Sucu_2025} further 
demonstrate that the addition of contextual information substantially improves stance detection accuracy.

We extend this line of work by jointly enforcing a structured output 
schema and applying DPO refinement \cite{wang2024selfdpo} with 
automatically generated preference pairs to reduce false negatives, 
combining the benefits of structured fine-tuning and 
preference-based optimization.

\section{Method}
\subsection{Problem setup}
We address Subtask~1 (Polarization Detection) of SemEval-2026 Task~9 (POLAR) \cite{naseem2026polarbenchmarkmultilingualmulticultural}, formulated as binary classification of social media posts as polarized ($y{=}1$) or not ($y{=}0$). In addition to the label, we generate a structured rationale: a target group, claim type, and a 6-category manifestation checklist (Stereotype, Vilification, Dehumanization, Extreme Language, Lack of Empathy, Invalidation) adopted from the organizers' Subtask~3 scheme \cite{polar2026taskwebsite}, plus a free-form justification.

\subsection{Approach}

\begin{figure}[h]
\centering
\resizebox{0.9\linewidth}{!}{ % Adjust 0.9 to 1.0 to fill the whole width
\begin{tikzpicture}[
    node distance=8mm and 8mm,
    box/.style={
        rectangle, 
        draw, 
        rounded corners=1mm, 
        text width=22mm, 
        minimum height=10mm, 
        align=center, 
        font=\small, 
        fill=blue!10
    },
    finalbox/.style={box, fill=green!15},
    arrow/.style={->, thick, >=stealth}
]

% First Row (Right to Left)
\node[box] (input) {Input Text};
\node[box, left=of input] (sft) {Structured SFT};
\node[box, left=of sft] (gen) {Generate Outputs};

% Second Row (Left to Right to follow a "U" flow, or Right to Left)
% Setting the second row to align under the first
\node[box, below=12mm of gen] (pref) {Preference Pairs};
\node[box, right=of pref] (dpo) {DPO Refinement};
\node[finalbox, right=of dpo] (final) {Final Prediction};

% Connections
\draw[arrow] (input) -- (sft);
\draw[arrow] (sft) -- (gen);

% Vertical transition
\draw[arrow] (gen.south) -- (pref.north);

\draw[arrow] (pref) -- (dpo);
\draw[arrow] (dpo) -- (final);

\end{tikzpicture}
}
\caption{Overview of our two-stage pipeline for Subtask~1 polarization detection: structured supervised fine-tuning (SFT) first generates schema-consistent outputs, and DPO refinement then learns from preference pairs to produce the final prediction.}
\label{fig:method-pipeline}
\end{figure}

We treat polarization detection as a generative task in which the model outputs both the label and a structured rationale. 
Our approach consists of a two-stage training pipeline: (i) structured supervised fine-tuning (SFT) to obtain a format-following polarization detector, and (ii) Direct Preference Optimization (DPO) refinement to improve decision quality, with a focus on reducing costly false negatives, while preserving the same output schema.

\paragraph{Stage 1: Structured supervised fine-tuning (SFT).}
We fine-tune an instruction-tuned LLM to emit a single structured record that includes the binary label and a rationale, enforcing a fixed output schema for consistency.

\paragraph{Stage 2: Preference-based refinement with DPO.}
Starting from the SFT checkpoint, we apply Direct Preference Optimization (DPO) \cite{rafailov2024directpreferenceoptimizationlanguage} to refine the decision boundary under the same output schema.

\textbf{Preference pair construction.} We build pairs by contrasting model outputs that (i) correctly classified content versus (ii) plausible but overly conservative outputs that wrongly label the instance. We employ categories obtained from the Subtask~3 manifestation indicators to prioritize outputs that recognize explicit vilification, dehumanization, or extreme language.

\section{Experiments}
\subsection{Experimental setup}

\paragraph{Dataset}
We evaluate our approach on the POLAR @ SemEval-2026 dataset released by the task organizers \cite{naseem2026polarbenchmarkmultilingualmulticultural,polar2026taskwebsite}.
Although the full dataset spans 22 languages, this work focuses exclusively on the English subset; the non-English portions of the dataset were not processed using the method described here, and our final task submission covered English only.
The \textit{train} and \textit{validation} splits were released before evaluation. 
We use these official train and validation splits as-is and report the English subset statistics in Table~\ref{tab:english-dataset-stats}.

\begin{table}[t]
\centering
\small
\begin{tabular}{lccc}
\hline
Split & Total & Polarized & Non-polarized \\ 
\hline
Train & 3,222 & 1,175 (36.5\%) & 2,047 (63.5\%) \\ 
Dev & 160 & 59 (36.9\%) & 101 (63.1\%) \\ 
Test & 1,452 & 533 (36.7\%) & 919 (63.3\%) \\ 
\hline
Total & 4,834 & 1,767 (36.6\%) & 3,067 (63.4\%) \\ 
\hline
\end{tabular}
\caption{English dataset statistics and label distribution across official splits.}
\label{tab:english-dataset-stats}
\end{table}

\paragraph{Baseline fine-tuning pipeline.}
We establish a simple fine-tuning baseline by splitting the data into train/validation/test (80/10/10), fine-tuning a classifier with LoRA adapters, and evaluating with macro precision, recall, and F1.

\subsubsection{Stage 1: Structured Supervised Fine-Tuning}

\textbf{Base Model} We initialize with the Qwen 2.5-7B-Instruct model \cite{qwen25}, a decoder-only large language model. Structured SFT training data was generated using Gemma 3 27B \cite{gemmateam2025gemma3technicalreport} served via Ollama, using the slot-filling prompt template described in Appendix~\ref{sec:sft-prompt}.

\textbf{Parameter-Efficient Fine-Tuning}
We fine-tune with LoRA \cite{hu2021loralowrankadaptationlarge} on Qwen/Qwen2.5-7B-Instruct using attention projections only (full hyperparameters in Appendix Table~\ref{tab:sft-hparams}).

To address class imbalance, we also test with a weighted loss upweighting polarized examples with class weights [1.0, 1.5].

\textbf{Generation Protocol}
Each prompt follows the format below; the model generates a structured rationale followed by a binary label, which is extracted via regex.
\begin{quote}
\texttt{Input: \{text\}}\\
\texttt{Reasoning:}
\end{quote}

\subsubsection{Stage 2: Preference-based refinement with DPO}
We apply DPO to refine the SFT model by learning from preference pairs that distinguish higher-quality reasoning and correct labels from weaker or incorrect outputs. We specifically try to address SFT’s tendency to produce false negatives.

DPO optimizes the model to prefer chosen responses over rejected ones via a contrastive loss, implicitly learning a reward signal without requiring a separate reward model \cite{rafailov2024directpreferenceoptimizationlanguage}. Compared to other RLHF techniques such as GRPO \cite{shao2024deepseekmath}, DPO is simpler to implement, more stable, and computationally lightweight \cite{rafailov2024directpreferenceoptimizationlanguage}.

\paragraph{Preference Pair Creation.}\label{sec:dpo-preference-pairs}
For each input, we sample multiple SFT completions at varied decoding temperatures to increase output diversity, then assign each completion one of three labels.
When an input yields heterogeneous labels, we construct preference pairs by ranking completions according to the labels above:
\[
\mathrm{CORRECT} \succ \mathrm{FP} \succ \mathrm{FN},
\]
treating higher-ranked completions as \emph{chosen} and lower-ranked ones as \emph{rejected}. We rank false negatives below false positives because missed polarized content poses greater harm in moderation contexts than over-flagging: an undetected polarizing post may spread unchecked, whereas a false positive can be reviewed and reversed. This ordering prioritizes recall recovery: SFT alone captures only half of polarized instances, making false negatives the dominant failure mode to correct.

After the release of the leaderboards, we also test a new pair-generation variant that uses two prompt settings per input rather than a single prompt: one that encourages a prediction of 1 and another that encourages a prediction of 0. The resulting outputs are then processed with the same ranking rule to determine \emph{chosen} and \emph{rejected} responses.

\paragraph{Training configuration.}
We report full DPO hyperparameters in Appendix Table~\ref{tab:dpo-hparams}.

\section{Results}
\label{sec:results}
Tables~\ref{tab:results-english} and~\ref{tab:results-english-metrics} report results on the \textbf{English development set} using the 80/10/10 split described above. 

\begin{table}[t]
\centering
\small
\begin{tabular}{lc}
\hline
Method & English Dev F1 \\
\hline
Zero-shot baseline & 0.7105 \\
SLMs & 0.7149 \\
SFT & 0.738 \\
SFT + DPO & 0.7893 \\
\hline
\end{tabular}
\caption{F1 scores on the English development set. SLMs = Small Language Models baseline (DistilBERT fine-tuned for sequence classification). Weighted loss during SFT did not improve performance.}
\label{tab:results-english}
\end{table}

Tables~\ref{tab:results-english} and ~\ref{tab:results-english-metrics} summarizes our English development results. As a lightweight baseline, we fine-tune DistilBERT \cite{sanh2019distilbert} for binary sequence classification (referred to as SLMs in Table~\ref{tab:results-english}); this serves as a reference point for the cost of supervised adaptation without instruction tuning or structured rationale generation. The zero-shot baseline starts at 0.7105 F1 on the dev set. The DistilBERT baseline reaches 0.7149, marginally above zero-shot. SFT further improves performance to 0.738, while DPO yields the strongest dev performance at 0.7893. 

On the English test set provided by the organizers, the SFT + DPO model reaches 0.7664 F1, indicating that preference refinement improves generalization beyond the development set. 

Weighted-loss supervised fine-tuning (SFT) did not outperform standard SFT in our experiments, suggesting that class imbalance was not the primary bottleneck. We therefore report unweighted SFT as the primary baseline.

\begin{table}[t]
\centering
\small
\begin{tabular}{lccc}
\hline
System & Rank (out of 60) & F1 \\ 
\hline
Highest-ranked system & 1 & 0.8252 \\ 
POLAR baseline & 47 & 0.7802 \\ 
Our system & 52 & 0.7664 \\ 
\hline
\end{tabular}
\caption{Unofficial English Subtask~1 leaderboard comparison (60 teams).}
\label{tab:leaderboard-rank}
\end{table}

Table~\ref{tab:leaderboard-rank} presents the ranking comparison between our system, the POLAR baseline, and the highest-ranked system.

\begin{table}[t]
\centering
\small
\begin{tabular}{lccc}
\hline
Metric & SFT & SFT + DPO \\
\hline
Accuracy & 0.7812 & 0.8000 \\
Precision & 0.8333 & 0.7077 \\
Recall & 0.5085 & 0.7797 \\
F1 (Binary) & 0.6316 & 0.7419 \\
F1 (Macro) & 0.7380 & 0.7893 \\
F1 (Micro) & 0.7812 & 0.8000 \\
\hline
\end{tabular}
\caption{English development set metrics comparing SFT vs. SFT + DPO.}
\label{tab:results-english-metrics}
\end{table}

Table~\ref{tab:results-english-metrics} highlights that DPO substantially increases recall (0.5085 $\rightarrow$ 0.7797), which reduces false negatives, but this comes with a drop in precision (0.8333 $\rightarrow$ 0.7077). We can explain this pattern by considering that DPO shifts the decision boundary toward preferring polarized outputs in borderline cases, increasing sensitivity at the cost of more false positives. The corresponding gains in F1 (binary and macro) indicate improved sensitivity to polarized content.

A qualitative example showing DPO correcting a false negative is provided in Appendix Table~\ref{tab:qual-example}.

\section{Experiments Post CodaBench Submission}

After the release of the official rankings and leaderboard on  GitHub, we trained both SFT and SFT + DPO on the full training  set. For the DPO runs, we also experimented with  LLM-as-a-judge filtering of preference pairs using  DeepSeek-R1 \cite{deepseekai2025deepseekr1},  following the same filtering procedure described in  Section~\ref{sec:post-submission}. Full details in \ref{sec:pair-generation}.

Tables~\ref{tab:rationale-ablation}  and~\ref{tab:dpo-pair-size} report results on the \textbf{official  English test set} ($n{=}1{,}452$), using models trained on the  full official training split. 
When trained on all 3,222 examples from the official English training split, SFT achieved an F1 score of $0.7712$ after 3 epochs and $0.7795$ after 10 epochs on the test set. Applying DPO to the best-performing SFT model resulted in an F1 score of $0.7889$. 

Training data was re-validated by Claude~3.5 Sonnet (hereafter \textit{Rejudged Sonnet} dataset), and DPO preference pairs were filtered using DeepSeek-R1 \cite{deepseekai2025deepseekr1} as an LLM judge, yielding a balanced 62:38 FP:FN ratio. 

\paragraph{Rejudged Sonnet dataset.}
The Rejudged Sonnet dataset is derived from the official English training split (3,222 examples) by re-validating every label with Claude~3.5 Sonnet as an LLM judge. Reasoning for each example was generated by GPT OSS 120B Nitro; Claude~3.5 Sonnet then evaluated the generated reasoning and revised labels where the reasoning did not support the original annotation. Each label was then also manually reviewed. Of the 3,178 training examples with an exact text match in the rejudged set, 196 (6.2\%) received a revised label, finally producing a net increase in the proportion of polarized examples.

\begin{table}[h]
\centering
\small
\setlength{\tabcolsep}{4pt}
\begin{tabular}{llcccc}
\hline
\textbf{Model} & \textbf{mF1} & \textbf{Acc.} & \textbf{P(1)} & \textbf{R(1)} \\
\hline
Mistral-Nemo  & 0.7963 & 0.8030 & 0.7928 & 0.8121 \\
Qwen2.5-7B         & 0.7835 & 0.7899 & 0.7811 & 0.8006 \\
\hline
\end{tabular}
\caption{Label-only SFT performance on the English test set ($n=1{,}452$)
trained on the original unmodified labels without the Rejudged Sonnet dataset.} 
\label{tab:original-label-baseline}
\end{table}

\subsection{DPO preference pair data}
Our DPO dataset is built automatically from SFT generations and is designed to preserve the same structured response format used during supervised fine-tuning. For details on how these pairs are constructed, please see Section~\ref{sec:dpo-preference-pairs}.

The following analysis was conducted after the task ranking was released and was not used to generate the leaderboard predictions.

Using the two-prompt method at varied temperatures, we generate 721 candidate pairs, filtered to 330 by deduplication and length ratio (Table~\ref{tab:dpo-pair-size}). More pairs reduce FNs but degrade overall F1 due to noise, confirming that LLM-as-a-judge quality control \cite{yu2025improvellmasajudgeabilitygeneral} is needed before training.

Preference pairs for post-submission DPO training on Mistral-Nemo were constructed exclusively from inputs where the Mistral-Nemo SFT checkpoint produced incorrect outputs. For each such input, the two-prompt method was applied using Llama-3.3-70B \cite{grattafiori2024llama3herdmodels}, yielding a pool of candidate responses. These candidates were ranked according to the ordering CORRECT $\succ$ FP $\succ$ FN.

The resulting pairs were then evaluated by DeepSeek-R1 \cite{deepseekai2025deepseekr1}, acting as an LLM-based judge, to remove low-confidence or inconsistent comparisons. This filtering step produced a final dataset of 299 preference pairs, with a 62:38 ratio of FP to FN cases. Full construction details and prompts are provided in Appendix~\ref{sec:pair-generation}.

\begin{table}[t]
\centering
\small
\begin{tabular}{lccc}
\hline
Mode & F1 & FNs & FPs \\
\hline
SFT & 0.7795 & 158 & 137 \\
SFT with DPO (330 pairs) & 0.7889 & 132 & 155 \\
SFT with DPO (721 pairs) & 0.7637 & 64 & 274 \\
\hline
\end{tabular}
\caption{Effect of DPO preference pair count on F1, false negatives (FNs), and false positives (FPs). SFT baseline is the 10-epoch full-training-set model; development-set results appear in Tables~\ref{tab:results-english} and~\ref{tab:results-english-metrics}.}
\label{tab:dpo-pair-size}
\end{table}

\subsection{Structured Rationale Ablation}
\label{sec:rationale-ablation}

To assess whether the structured rationale contributes to classification performance, we compare four conditions in a controlled ablation on the English test set ($n=1{,}452$): label-only SFT, label-only SFT followed by our reasoning DPO setup (label-only+DPO), reasoning SFT, and reasoning SFT+DPO. All SFT conditions use 3 epochs, the same base model, LoRA configuration, and training data.

\begin{table}[t]
\centering
\footnotesize
\setlength{\tabcolsep}{3pt}
\begin{tabular}{lcccc}
\hline
Condition & Acc. & P(1) & R(1) & mF1 \\
\hline
Label-only SFT       & 0.792 & 0.777 & 0.7884 & 0.781 \\
Label-only + DPO     & 0.720 & 0.618 & 0.625 & 0.699 \\
Reasoning SFT        & 0.793 & 0.745 & 0.662 & 0.771 \\
Reasoning + DPO      & 0.802 & 0.732 & 0.704 & 0.789 \\
\hline
\end{tabular}
\caption{Structured rationale ablation on the English test set. P(1) and R(1) = precision and recall for the polarized class; mF1 = Macro-F1. Best in bold.}
\label{tab:rationale-ablation}
\end{table}

Table~\ref{tab:rationale-ablation} shows that label-only SFT outperforms reasoning SFT in isolation (Macro-F1: 0.7811 vs.\ 0.771), consistent with findings that SFT on full chain-of-thought sequences dilutes gradient signal on the final label token \cite{shi2025rethinkingsupervisedfinetuningemphasizing}. However, label-only SFT+DPO collapses to 0.699, while reasoning SFT+DPO achieves the best result (0.789), showing that the rationale's value is in enabling DPO rather than improving the SFT decision boundary. These ablations use Qwen2.5-7B-Instruct; post-submission results (Section~\ref{sec:post-submission}) show that reasoning data does benefit the larger Mistral-Nemo model.

\subsection{Post-submission Results}
\label{sec:post-submission}

All experiments here were conducted after the official submission deadline and were not submitted to CodaBench. A full beta sweep and additional analysis are in Appendix~\ref{sec:post-submission-detail}.

\begin{table}[t]
\centering
\footnotesize
\setlength{\tabcolsep}{3pt}
\begin{tabular}{lc}
\hline
System & mF1 \\
\hline
Qwen2.5-7B Label-Only SFT (Rejudged) & 0.7811 \\
Mistral-Nemo Label-Only SFT (Rejudged) & 0.8019 \\
Mistral-Nemo SFT (Rejudged) & 0.8097 \\
Mistral-Nemo DPO (FP:FN 62:38, $\beta$=0.3) & 0.8162 \\
\hline
\end{tabular}
\caption{Post-submission results on the English test set ($n{=}1{,}452$). None submitted to CodaBench. mF1 = Macro-F1.}
\label{tab:post-submission}
\end{table}

Post-submission experiments replaced Qwen2.5-7B-Instruct with Mistral-Nemo-Instruct-2407 \cite{mistralnemo2024}, a 12B model with strong multilingual instruction-following. 

Table~\ref{tab:post-submission} shows that Mistral-Nemo SFT with Rejudged Sonnet data (0.8097) surpasses both the organiser baseline (0.7802) and the label-only variant (0.8019), confirming that structured reasoning data provides a meaningful training signal for the larger model. Applying quality-filtered DPO at $\beta$=0.3 further improves to 0.8162 \cite{polar2026leaderboard}, up from 0.7664 (rank~52) for the submitted system.

Results of the beta sweep over nine $\beta$ values (0.1--0.5) are reported in Appendix Table~\ref{tab:beta-sweep}.

\section{Conclusion}
We presented a two-stage system for polarization detection combining structured SFT with DPO refinement. The submitted system, based on Qwen2.5-7B-Instruct, achieved 0.7664 Macro-F1 on the English test set.  

Post-submission improvements (Section~\ref{sec:post-submission}) raise performance to 0.8162 Macro-F1: surpassing the organiser baseline (0.7802) on the unofficial English leaderboard  \cite{polar2026leaderboard}. 

\section*{Limitations}
We observed a subset of ambiguous examples in the training dataset where cues were mixed or context-dependent, and annotator intent was not always clear from the text alone. These cases were especially challenging for both SFT and DPO, leading to inconsistent predictions. The structured rationale does not fully resolve this ambiguity.

A limitation of our submitted system is that, for Qwen2.5-7B-Instruct, structured rationales do not outperform a label-only baseline. Label-only SFT achieves 0.7811 Macro-F1 versus 0.771 for reasoning SFT. At this scale, rationales mainly recover performance lost during reasoning-based fine-tuning via preference optimization, rather than improving it. Post-submission results indicate this is a scale effect, with clearer gains from structured reasoning emerging at 12B parameters.

The post-submission experiments use a re-validated training set that differs from the data used for the official submission; improvements over the submitted system therefore reflect both the stronger base model and the change in training labels.

Precision remains relatively low across all configurations using rationale-based SFT (best: 0.76), reflecting the subtle boundary between emphatic-but-neutral language and genuinely polarizing content, as well as the model's tendency toward high recall.

\section*{Future Work}
Post-submission results confirm that preference pair quality is the primary  DPO bottleneck. Future work should investigate loss masking  \cite{shi2025rethinkingsupervisedfinetuningemphasizing} to recover the recall cost of reasoning SFT,  and explore SimPO \cite{meng2024simpo} or KTO \cite{ethayarajh2024kto} as  more stable preference optimisation alternatives. The precision-recall  trade-off and the causes of DPO instability in smaller models warrant  further study --- in particular, whether reference-free objectives can  stabilise DPO for 7B-scale models. Extending the pipeline to non-English  languages in the POLAR benchmark \cite{naseem2026polarbenchmarkmultilingualmulticultural}  is a direct next step, as is evaluating whether the Rejudged Sonnet quality  improvement transfers to other base models. Generating reasoning chains using  a stronger judge model (e.g. GPT-4o or Claude~3.7  Sonnet) could yield higher-quality training data  and further improve both SFT and DPO performance. Sourcing additional training  data from related polarization and hate-speech datasets \cite{basile-etal-2019-semeval}  would also increase coverage of underrepresented manifestation types such as  subtle Invalidation.

% If you didn't upload the Anthology .bib/.bst files, switch to a standard BibTeX style (e.g., plainnat)
% or upload: anthology.bib, custom.bib, and acl_natbib.bst.
\bibliography{custom}

@misc{yu2025improvellmasajudgeabilitygeneral,
      title={Improve LLM-as-a-Judge Ability as a General Ability}, 
      author={Jiachen Yu and Shaoning Sun and Xiaohui Hu and Jiaxu Yan and Kaidong Yu and Xuelong Li},
      year={2025},
      eprint={2502.11689},
      archivePrefix={arXiv},
      primaryClass={cs.CL},
      url={https://arxiv.org/abs/2502.11689}, 
}

@article{rafailov2024directpreferenceoptimizationlanguage,
  title   = {Direct Preference Optimization: Your Language Model is Secretly a Reward Model},
  author  = {Rafailov, Rafael and Sharma, Archit and Mitchell, Eric and Ermon, Stefano and Manning, Christopher D. and Finn, Chelsea},
  year    = {2024},
  journal = {arXiv preprint arXiv:2305.18290},
}

@misc{naseem2026polarbenchmarkmultilingualmulticultural,
      title={POLAR: A Benchmark for Multilingual, Multicultural, and Multi-Event Online Polarization},
      author={Usman Naseem and Robert Geislinger and Juan Ren and Sarah Kohail and Rudy Garrido Veliz and P Sam Sahil and Yiran Zhang and Marco Antonio Stranisci and Idris Abdulmumin and Özge Alacam and Cengiz Acartürk and Aisha Jabr and Saba Anwar and Abinew Ali Ayele and Simona Frenda and Alessandra Teresa Cignarella and Elena Tutubalina and Oleg Rogov and Aung Kyaw Htet and Xintong Wang and Surendrabikram Thapa and Kritesh Rauniyar and Tanmoy Chakraborty and Arfeen Zeeshan and Dheeraj Kodati and Satya Keerthi and Sahar Moradizeyveh and Firoj Alam and Arid Hasan and Syed Ishtiaque Ahmed and Ye Kyaw Thu and Shantipriya Parida and Ihsan Ayyub Qazi and Lilian Wanzare and Nelson Odhiambo Onyango and Clemencia Siro and Jane Wanjiru Kimani and Ibrahim Said Ahmad and Adem Chanie Ali and Martin Semmann and Chris Biemann and Shamsuddeen Hassan Muhammad and Seid Muhie Yimam},
      year={2026},
      eprint={2505.20624},
      archivePrefix={arXiv},
      primaryClass={cs.CL},
      url={https://arxiv.org/abs/2505.20624},
}

@inproceedings{naseem-etal-2026-polar,
    title     = {{SemEval-2026 Task 9: Detecting Multilingual, Multicultural and Multievent Online Polarization}},
    author    = {Naseem, Usman and Geislinger, Robert and Ren, Juan and Kohail, Sarah and Garrido Veliz, Rudy and Sam Sahil, P and Zhang, Yiran and Stranisci, Marco Antonio and Abdulmumin, Idris and Alacam, Özge and Acarürk, Cengiz and Jabr, Aisha and Anwar, Saba and Ayele, Abinew Ali and Tutubalina, Elena and Htet, Aung Kyaw and Wang, Xintong and Thapa, Surendrabikram and Chakraborty, Tanmoy and Kodati, Dheeraj and Moradizeyveh, Sahar and Alam, Firoj and Thu, Ye Kyaw and Parida, Shantipriya and Qazi, Ihsan Ayyub and Onyango, Nelson Odhiambo and Siro, Clemencia and Ahmad, Ibrahim Said and Wanzare, Lilian and Ali, Adem Chanie and Semmann, Martin and Biemann, Chris and Muhammad, Shamsuddeen Hassan and Yimam, Seid Muhie},
    booktitle = {{Proceedings of the 20th International Workshop on Semantic Evaluation (SemEval-2026)}},
    year      = {2026},
    publisher = {{Association for Computational Linguistics}},
    address = {{San Diego, CA, USA}}
}

@misc{qwen25,
    title = {Qwen2.5: A Party of Foundation Models},
    url = {https://qwenlm.github.io/blog/qwen2.5/},
    author = {Qwen Team},
    month = {September},
    year = {2024}
}

@misc{maggini2025llmshyperpartisanfakepolarized,
      title={Are LLMs Enough for Hyperpartisan, Fake, Polarized and Harmful Content Detection? Evaluating In-Context Learning vs. Fine-Tuning}, 
      author={Michele Joshua Maggini and Dhia Merzougui and Rabiraj Bandyopadhyay and Gaël Dias and Fabrice Maurel and Pablo Gamallo},
      year={2025},
      eprint={2509.07768},
      archivePrefix={arXiv},
      primaryClass={cs.CL},
      url={https://arxiv.org/abs/2509.07768}, 
}

@inproceedings{basile-etal-2019-semeval,
    title = "{S}em{E}val-2019 Task 5: Multilingual Detection of Hate Speech Against Immigrants and Women in {T}witter",
    author = "Basile, Valerio  and
      Bosco, Cristina  and
      Fersini, Elisabetta  and
      Nozza, Debora  and
      Patti, Viviana  and
      Rangel Pardo, Francisco Manuel  and
      Rosso, Paolo  and
      Sanguinetti, Manuela",
    editor = "May, Jonathan  and
      Shutova, Ekaterina  and
      Herbelot, Aurelie  and
      Zhu, Xiaodan  and
      Apidianaki, Marianna  and
      Mohammad, Saif M.",
    booktitle = "Proceedings of the 13th International Workshop on Semantic Evaluation",
    month = jun,
    year = "2019",
    address = "Minneapolis, Minnesota, USA",
    publisher = "Association for Computational Linguistics",
    url = "https://aclanthology.org/S19-2007/",
    doi = "10.18653/v1/S19-2007",
    pages = "54--63"
}

@misc{hu2021loralowrankadaptationlarge,
      title={LoRA: Low-Rank Adaptation of Large Language Models}, 
      author={Edward J. Hu and Yelong Shen and Phillip Wallis and Zeyuan Allen-Zhu and Yuanzhi Li and Shean Wang and Lu Wang and Weizhu Chen},
      year={2021},
      eprint={2106.09685},
      archivePrefix={arXiv},
      primaryClass={cs.CL},
      url={https://arxiv.org/abs/2106.09685}, 
}

@misc{polar2026taskwebsite,
      title={POLAR @ SemEval-2026 Task 9: Detecting Multilingual, Multicultural and Multievent Online Polarization},
      author={{POLAR Task Organizers}},
      year={2026},
      url={https://polar-semeval.github.io/},
      note={Accessed: 2026-02-27},
}

@misc{pangtey2025llmsstancedetectionsurvey,
      title={Large Language Models Meet Stance Detection: A Survey of Tasks, Methods, Applications, Challenges and Future Directions},
      author={Lata Pangtey and Anukriti Bhatnagar and Shubhi Bansal and Shahid Shafi Dar and Nagendra Kumar},
      year={2025},
      eprint={2505.08464},
      archivePrefix={arXiv},
      primaryClass={cs.CL},
      doi={10.48550/arXiv.2505.08464},
      url={https://arxiv.org/abs/2505.08464},
}

@misc{gemmateam2025gemma3technicalreport,
      title={Gemma 3 Technical Report}, 
      author={{Gemma Team} and Aishwarya Kamath and Johan Ferret and Shreya Pathak and Nino Vieillard and Ramona Merhej and Sarah Perrin and Tatiana Matejovicova and Alexandre Ramé and Morgane Rivière and Louis Rouillard and Thomas Mesnard and Geoffrey Cideron and Jean-bastien Grill and Sabela Ramos and Edouard Yvinec and Michelle Casbon and Etienne Pot and Ivo Penchev and Gaël Liu and Francesco Visin and Kathleen Kenealy and Lucas Beyer and Xiaohai Zhai and Anton Tsitsulin and Robert Busa-Fekete and Alex Feng and Noveen Sachdeva and Benjamin Coleman and Yi Gao and Basil Mustafa and Iain Barr and Emilio Parisotto and David Tian and Matan Eyal and Colin Cherry and Jan-Thorsten Peter and Danila Sinopalnikov and Surya Bhupatiraju and Rishabh Agarwal and Mehran Kazemi and Dan Malkin and Ravin Kumar and David Vilar and Idan Brusilovsky and Jiaming Luo and Andreas Steiner and Abe Friesen and Abhanshu Sharma and Abheesht Sharma and Adi Mayrav Gilady and Adrian Goedeckemeyer and Alaa Saade and Alex Feng and Alexander Kolesnikov and Alexei Bendebury and Alvin Abdagic and Amit Vadi and András György and André Susano Pinto and Anil Das and Ankur Bapna and Antoine Miech and Antoine Yang and Antonia Paterson and Ashish Shenoy and Ayan Chakrabarti and Bilal Piot and Bo Wu and Bobak Shahriari and Bryce Petrini and Charlie Chen and Charline Le Lan and Christopher A. Choquette-Choo and CJ Carey and Cormac Brick and Daniel Deutsch and Danielle Eisenbud and Dee Cattle and Derek Cheng and Dimitris Paparas and Divyashree Shivakumar Sreepathihalli and Doug Reid and Dustin Tran and Dustin Zelle and Eric Noland and Erwin Huizenga and Eugene Kharitonov and Frederick Liu and Gagik Amirkhanyan and Glenn Cameron and Hadi Hashemi and Hanna Klimczak-Plucińska and Harman Singh and Harsh Mehta and Harshal Tushar Lehri and Hussein Hazimeh and Ian Ballantyne and Idan Szpektor and Ivan Nardini and Jean Pouget-Abadie and Jetha Chan and Joe Stanton and John Wieting and Jonathan Lai and Jordi Orbay and Joseph Fernandez and Josh Newlan and Ju-yeong Ji and Jyotinder Singh and Kat Black and Kathy Yu and Kevin Hui and Kiran Vodrahalli and Klaus Greff and Linhai Qiu and Marcella Valentine and Marina Coelho and Marvin Ritter and Matt Hoffman and Matthew Watson and Mayank Chaturvedi and Michael Moynihan and Min Ma and Nabila Babar and Natasha Noy and Nathan Byrd and Nick Roy and Nikola Momchev and Nilay Chauhan and Noveen Sachdeva and Oskar Bunyan and Pankil Botarda and Paul Caron and Paul Kishan Rubenstein and Phil Culliton and Philipp Schmid and Pier Giuseppe Sessa and Pingmei Xu and Piotr Stanczyk and Pouya Tafti and Rakesh Shivanna and Renjie Wu and Renke Pan and Reza Rokni and Rob Willoughby and Rohith Vallu and Ryan Mullins and Sammy Jerome and Sara Smoot and Sertan Girgin and Shariq Iqbal and Shashir Reddy and Shruti Sheth and Siim Põder and Sijal Bhatnagar and Sindhu Raghuram Panyam and Sivan Eiger and Susan Zhang and Tianqi Liu and Trevor Yacovone and Tyler Liechty and Uday Kalra and Utku Evci and Vedant Misra and Vincent Roseberry and Vlad Feinberg and Vlad Kolesnikov and Woohyun Han and Woosuk Kwon and Xi Chen and Yinlam Chow and Yuvein Zhu and Zichuan Wei and Zoltan Egyed and Victor Cotruta and Minh Giang and Phoebe Kirk and Anand Rao and Kat Black and Nabila Babar and Jessica Lo and Erica Moreira and Luiz Gustavo Martins and Omar Sanseviero and Lucas Gonzalez and Zach Gleicher and Tris Warkentin and Vahab Mirrokni and Evan Senter and Eli Collins and Joelle Barral and Zoubin Ghahramani and Raia Hadsell and Yossi Matias and D. Sculley and Slav Petrov and Noah Fiedel and Noam Shazeer and Oriol Vinyals and Jeff Dean and Demis Hassabis and Koray Kavukcuoglu and Clement Farabet and Elena Buchatskaya and Jean-Baptiste Alayrac and Rohan Anil and Dmitry and Lepikhin and Sebastian Borgeaud and Olivier Bachem and Armand Joulin and Alek Andreev and Cassidy Hardin and Robert Dadashi and Léonard Hussenot},
      year={2025},
      eprint={2503.19786},
      archivePrefix={arXiv},
      primaryClass={cs.CL},
      url={https://arxiv.org/abs/2503.19786}, 
}

@inproceedings{gunel2021scl,
  title     = {Supervised Contrastive Learning for Pre-trained Language Model Fine-tuning},
  author    = {Gunel, Beliz and Du, Jingfei and Conneau, Alexis and Stoyanov, Veselin},
  booktitle = {International Conference on Learning Representations (ICLR)},
  year      = {2021},
  url       = {https://openreview.net/forum?id=cu7IUiOhujH}
}

@article{Loru_2025,
   title={Ideology and polarization set the agenda on social media},
   volume={15},
   ISSN={2045-2322},
   url={http://dx.doi.org/10.1038/s41598-025-19776-z},
   DOI={10.1038/s41598-025-19776-z},
   number={1},
   journal={Scientific Reports},
   publisher={Springer Science and Business Media LLC},
   author={Loru, Edoardo and Galeazzi, Alessandro and Bonetti, Anita and Sangiorgio, Emanuele and Di Marco, Niccolò and Cinelli, Matteo and Falkenberg, Max and Baronchelli, Andrea and Quattrociocchi, Walter},
   year={2025},
   month=oct }

@inproceedings{Sucu_2025,
   title={Exploiting contextual information to improve stance detection in informal political discourse with LLMs},
   url={http://dx.doi.org/10.18653/v1/2025.acl-srw.86},
   DOI={10.18653/v1/2025.acl-srw.86},
   booktitle={Proceedings of the 63rd Annual Meeting of the Association for Computational Linguistics (Volume 4: Student Research Workshop)},
   publisher={Association for Computational Linguistics},
   author={Sucu, Arman Engin and Zhou, Yixiang and Nascimento, Mario A. and Mullen, Tony},
   year={2025},
   pages={1097–1110} }

@misc{bommasani2022opportunitiesrisksfoundationmodels,
      title={On the Opportunities and Risks of Foundation Models}, 
      author={Rishi Bommasani and Drew A. Hudson and Ehsan Adeli and Russ Altman and Simran Arora and Sydney von Arx and Michael S. Bernstein and Jeannette Bohg and Antoine Bosselut and Emma Brunskill and Erik Brynjolfsson and Shyamal Buch and Dallas Card and Rodrigo Castellon and Niladri Chatterji and Annie Chen and Kathleen Creel and Jared Quincy Davis and Dora Demszky and Chris Donahue and Moussa Doumbouya and Esin Durmus and Stefano Ermon and John Etchemendy and Kawin Ethayarajh and Li Fei-Fei and Chelsea Finn and Trevor Gale and Lauren Gillespie and Karan Goel and Noah Goodman and Shelby Grossman and Neel Guha and Tatsunori Hashimoto and Peter Henderson and John Hewitt and Daniel E. Ho and Jenny Hong and Kyle Hsu and Jing Huang and Thomas Icard and Saahil Jain and Dan Jurafsky and Pratyusha Kalluri and Siddharth Karamcheti and Geoff Keeling and Fereshte Khani and Omar Khattab and Pang Wei Koh and Mark Krass and Ranjay Krishna and Rohith Kuditipudi and Ananya Kumar and Faisal Ladhak and Mina Lee and Tony Lee and Jure Leskovec and Isabelle Levent and Xiang Lisa Li and Xuechen Li and Tengyu Ma and Ali Malik and Christopher D. Manning and Suvir Mirchandani and Eric Mitchell and Zanele Munyikwa and Suraj Nair and Avanika Narayan and Deepak Narayanan and Ben Newman and Allen Nie and Juan Carlos Niebles and Hamed Nilforoshan and Julian Nyarko and Giray Ogut and Laurel Orr and Isabel Papadimitriou and Joon Sung Park and Chris Piech and Eva Portelance and Christopher Potts and Aditi Raghunathan and Rob Reich and Hongyu Ren and Frieda Rong and Yusuf Roohani and Camilo Ruiz and Jack Ryan and Christopher Ré and Dorsa Sadigh and Shiori Sagawa and Keshav Santhanam and Andy Shih and Krishnan Srinivasan and Alex Tamkin and Rohan Taori and Armin W. Thomas and Florian Tramèr and Rose E. Wang and William Wang and Bohan Wu and Jiajun Wu and Yuhuai Wu and Sang Michael Xie and Michihiro Yasunaga and Jiaxuan You and Matei Zaharia and Michael Zhang and Tianyi Zhang and Xikun Zhang and Yuhui Zhang and Lucia Zheng and Kaitlyn Zhou and Percy Liang},
      year={2022},
      eprint={2108.07258},
      archivePrefix={arXiv},
      primaryClass={cs.LG},
      url={https://arxiv.org/abs/2108.07258}, 
}

@misc{devlin2019bertpretrainingdeepbidirectional,
      title={BERT: Pre-training of Deep Bidirectional Transformers for Language Understanding}, 
      author={Jacob Devlin and Ming-Wei Chang and Kenton Lee and Kristina Toutanova},
      year={2019},
      eprint={1810.04805},
      archivePrefix={arXiv},
      primaryClass={cs.CL},
      url={https://arxiv.org/abs/1810.04805}, 
}

@article{franceschelli2025creativity,
  title={On the creativity of large language models},
  author={Franceschelli, Giorgio and Musolesi, Mirco},
  journal={AI \& society},
  volume={40},
  number={5},
  pages={3785--3795},
  year={2025},
  publisher={Springer}
}

@article{li2024pre,
  title={Pre-trained language models for text generation: A survey},
  author={Li, Junyi and Tang, Tianyi and Zhao, Wayne Xin and Nie, Jian-Yun and Wen, Ji-Rong},
  journal={ACM Computing Surveys},
  volume={56},
  number={9},
  pages={1--39},
  year={2024},
  publisher={ACM New York, NY}
}

@inproceedings{zhang-etal-2024-sentiment,
    title = "Sentiment Analysis in the Era of Large Language Models: A Reality Check",
    author = "Zhang, Wenxuan  and
      Deng, Yue  and
      Liu, Bing  and
      Pan, Sinno  and
      Bing, Lidong",
    editor = "Duh, Kevin  and
      Gomez, Helena  and
      Bethard, Steven",
    booktitle = "Findings of the Association for Computational Linguistics: NAACL 2024",
    month = jun,
    year = "2024",
    address = "Mexico City, Mexico",
    publisher = "Association for Computational Linguistics",
    url = "https://aclanthology.org/2024.findings-naacl.246/",
    doi = "10.18653/v1/2024.findings-naacl.246",
    pages = "3881--3906"
}

@misc{mistralnemo2024,
  author       = {{Mistral AI}},
  title        = {Mistral NeMo},
  year         = {2024},
  howpublished = {\url{https://mistral.ai/news/mistral-nemo/}},
  note         = {Accessed: April 2026}
}

@misc{deepseekai2025deepseekr1,
      title={DeepSeek-R1: Incentivizing Reasoning Capability in 
      LLMs via Reinforcement Learning},
      author={{DeepSeek-AI}},
      year={2025},
      eprint={2501.12948},
      archivePrefix={arXiv},
      primaryClass={cs.CL},
      url={https://arxiv.org/abs/2501.12948},
}

@article{wei2022chain,
  title={Chain-of-thought prompting elicits reasoning in large language models},
  author={Wei, Jason and Wang, Xuezhi and Schuurmans, Dale and Bosma, Maarten and Ichter, Brian and Xia, Fei and Chi, Ed and Le, Quoc and Zhou, Denny},
  journal={Advances in Neural Information Processing Systems},
  volume={35},
  pages={24824--24837},
  year={2022}
}

@article{wang2024selfdpo,
  title={Self-Training with Direct Preference Optimization Improves Chain-of-Thought Reasoning},
  author={Wang, Tianduo and Li, Shichen and Lu, Wei},
  journal={arXiv preprint arXiv:2407.18248},
  year={2024}
}

@article{sanh2019distilbert,
  title={DistilBERT, a distilled version of BERT: smaller, faster, cheaper and lighter},
  author={Sanh, Victor and Debut, Lysandre and Chaumond, Julien and Wolf, Thomas},
  journal={arXiv preprint arXiv:1910.01108},
  year={2019}
}

@inproceedings{meng2024simpo,
  title     = {{SimPO}: Simple Preference Optimization with a Reference-Free Reward},
  author    = {Meng, Yu and Xia, Mengzhou and Chen, Danqi},
  booktitle = {Advances in Neural Information Processing Systems (NeurIPS)},
  year      = {2024},
  eprint    = {2405.14734},
  archivePrefix = {arXiv},
  primaryClass  = {cs.CL},
  url       = {https://arxiv.org/abs/2405.14734}
}

@inproceedings{ethayarajh2024kto,
  title     = {{KTO}: Model Alignment as Prospect Theoretic Optimization},
  author    = {Ethayarajh, Kawin and Xu, Winnie and Muennighoff, Niklas and 
               Jurafsky, Dan and Kiela, Douwe},
  booktitle = {Proceedings of the 41st International Conference on Machine Learning (ICML)},
  year      = {2024},
  eprint    = {2402.01306},
  archivePrefix = {arXiv},
  primaryClass  = {cs.LG},
  url       = {https://arxiv.org/abs/2402.01306}
}

@misc{shi2025rethinkingsupervisedfinetuningemphasizing,
      title={Rethinking Supervised Fine-Tuning: Emphasizing Key Answer Tokens for Improved LLM Accuracy}, 
      author={Xiaofeng Shi and Qian Kou and Yuduo Li and Hua Zhou},
      year={2025},
      eprint={2512.21017},
      archivePrefix={arXiv},
      primaryClass={cs.CL},
      url={https://arxiv.org/abs/2512.21017}, 
}

@misc{grattafiori2024llama3herdmodels,
  title={The Llama 3 Herd of Models},
  author={Grattafiori, Aaron and others},
  year={2024},
  eprint={2407.21783},
  archivePrefix={arXiv},
  primaryClass={cs.AI}
}

@misc{polar2026leaderboard,
  author       = {{Polar-SemEval}},
  title        = {{POLAR @ SemEval-2026 Leaderboards}},
  year         = {2026},
  howpublished = {\url{https://github.com/Polar-SemEval/Leaderboards}},
  note         = {Accessed: April 2026}
}

@article{shao2024deepseekmath,
  title={DeepSeekMath: Pushing the Limits of Mathematical Reasoning in Open Language Models},
  author={Shao, Zhihong and Wang, Peiyi and Zhu, Qihao and Xu, Runxin and Song, Junxian and Bi, Xiao and Zhang, Haowei and Zhang, Mingchuan and Li, Y.K. and Wu, Y. and others},
  journal={arXiv preprint arXiv:2402.03300},
  year={2024}
}

\FloatBarrier
\appendix
\section{Appendix}
\subsection{Post-submission Experimental Detail}
\label{sec:post-submission-detail}

\begin{figure}[H]
\centering
\resizebox{0.95\linewidth}{!}{
\begin{tikzpicture}[
    node distance=8mm and 6mm,
    box/.style={
        rectangle,
        draw,
        rounded corners=1mm,
        text width=24mm,
        minimum height=10mm,
        align=center,
        font=\small,
        fill=blue!10
    },
    databox/.style={box, fill=orange!15},
    judgebox/.style={box, fill=yellow!25},
    finalbox/.style={box, fill=green!15},
    arrow/.style={->, thick, >=stealth},
    darrow/.style={->, thick, >=stealth, dashed, gray}
]

% Row 1 (left to right): Training data → SFT Training → SFT Model
\node[databox] (data)             {Training Data\\(Rejudged Sonnet)};
\node[box, right=of data]  (sft)      {SFT Training};
\node[box, right=of sft]   (sftmodel) {SFT Model\\(Mistral-Nemo)};

% Row 2 (right to left): Varied sampling → Candidate Pairs → LLM Judge → Filtered Pairs
\node[box,      below=12mm of sftmodel] (sampling)  {Varied-temp\\Sampling};
\node[box,      left=of sampling]       (candpairs) {Candidate\\Pairs (Llama-3.3-70B)};
\node[judgebox, left=of candpairs]      (judge)     {LLM Judge\\(DeepSeek-R1)};
\node[box,      left=of judge]          (filtered)  {Filtered Pairs\\(62:38 FP:FN)};

% Row 3 (left to right): DPO Training → Final Model
\node[box,      below=12mm of filtered] (dpo)   {DPO Training\\($\beta$=0.3, 2 ep.)};
\node[finalbox, right=of dpo]           (final) {Final Model\\(\textbf{0.8162} F1)};

% Row 1 arrows
\draw[arrow] (data)     -- (sft);
\draw[arrow] (sft)      -- (sftmodel);

% SFT Model → Sampling (vertical transition)
\draw[arrow] (sftmodel.south) -- (sampling.north);

% Row 2 arrows
\draw[arrow] (sampling)  -- (candpairs);
\draw[arrow] (candpairs) -- (judge);
\draw[arrow] (judge)     -- (filtered);

% Filtered Pairs → DPO (vertical transition)
\draw[arrow] (filtered.south) -- (dpo.north);

% Row 3 arrow
\draw[arrow] (dpo) -- (final);

% Dashed: SFT Model → DPO as reference model (routes along right edge)
\draw[darrow] (sftmodel.east) -- ++(6mm,0) -- ++(0,-36mm) -| (dpo.east)
    node[midway, right, font=\scriptsize, gray] {reference model};

\end{tikzpicture}
}
\caption{Post-submission pipeline. Rejudged Sonnet training data produces an SFT checkpoint, which then generates completions at varied temperatures to form candidate preference pairs. DeepSeek-R1 filters these into a balanced set (62:38 FP:FN ratio). DPO then refines the same SFT checkpoint (dashed arrow) using the filtered pairs, producing the best post-submission result of 0.8162 Macro-F1. The Rejudged Sonnet training data was produced by generating 
reasoning with GPT OSS 120B Nitro and filtering labels with Claude~3.5 Sonnet as a judge (see Appendix~\ref{sec:pair-generation}).}
\label{fig:post-submission-pipeline}
\end{figure}

The post-submission setup is described in Section~\ref{sec:post-submission}. We provide stability and sensitivity details below. 

Across 20 DPO runs on Mistral-Nemo, using stale preference pairs, unfiltered pairs, or excessive training epochs consistently degraded performance. Epoch count is especially critical: 10 epochs on 260 pairs collapsed to Macro-F1 0.6508 vs.\ 0.8012 at 2 epochs.

A subsequent beta sweep over 9 values (0.1--0.5) on the same pair set confirms robustness: all configurations beat the SFT baseline (0.8097), with Macro-F1 ranging from 0.8065 to 0.8142 and no single beta dominating clearly (Table~\ref{tab:beta-sweep}). This flatness suggests pair quality and composition (FP:FN ratio) are the binding constraints, not $\beta$.

\begin{table}[h]
\centering
\small
\setlength{\tabcolsep}{4pt}
\begin{tabular}{ccccc}
\hline
$\beta$ & mF1 & Acc. & P(1) & R(1) \\
\hline
0.10 & 0.8127 & 0.8230 & 0.739 & 0.801 \\
0.15 & 0.8125 & 0.8230 & 0.740 & 0.799 \\
0.20 & 0.8136 & 0.8237 & 0.738 & 0.805 \\
0.25 & 0.8108 & 0.8209 & 0.734 & 0.803 \\
0.30 & \textbf{0.8162} & 0.8230 & 0.736 & 0.809 \\
0.35 & 0.8079 & 0.8182 & 0.731 & 0.799 \\
0.40 & 0.8071 & 0.8189 & 0.727 & 0.811 \\
0.45 & 0.8142 & 0.8258 & 0.737 & 0.816 \\
0.50 & 0.8065 & 0.8223 & 0.733 & 0.811 \\
\hline
SFT baseline & 0.8097 & 0.8182 & 0.719 & 0.827 \\
\hline
\end{tabular}
\caption{Beta sweep on 299 DeepSeek-R1-filtered DPO pairs (62:38 FP:FN ratio), Mistral-Nemo-Instruct-2407, 2 epochs. mF1 = Macro-F1; P(1)/R(1) = precision/recall for the polarized class.}
\label{tab:beta-sweep}
\end{table}

\subsection{Qualitative Example}
Table~\ref{tab:qual-example} shows a representative case from the English test set where SFT produced a false negative that DPO subsequently corrected. The SFT model marked all manifestation categories as absent and assigned label~0. After DPO refinement, the model identified \textit{Stereotype}, \textit{Vilification}, \textit{Lack of Empathy}, and \textit{Invalidation} as present, driving the correct label-1 prediction. This suggests that the structured manifestation checklist is doing real work in the DPO model: the label change is anchored to specific checklist fields rather than arising from an unconstrained shift in output distribution.

\begin{table}[h]
\centering
\small
\begin{tabular}{p{0.95\linewidth}}
\hline
\textbf{Input text} (\texttt{eng\_03a6df98}) \\
\hline
\textit{``Ending 2024 with an illadvised text conversation with my brother that ended with `Really little bro? I can deal with the atheism, but socialism?'''} \\
Gold label: 1 (Lack of Empathy: present; Invalidation: present) \\
\hline
\textbf{SFT output (label: 0 --- false negative)} \\
\hline
Target: specific individual \quad Claim type: evaluative opinion \\
Stereotype: absent \quad Vilification: absent \\
Dehumanization: absent \quad Extreme Language: absent \\
Lack of Empathy: \textbf{absent} \quad Invalidation: \textbf{absent} \\
Decision basis: The text is a personal, non-political anecdote about a family disagreement that does not contain any of the defined polarization manifestations. \\
Final Answer: \textbf{0} \\
\hline
\textbf{SFT + DPO output (label: 1 --- corrected)} \\
\hline
Target: \textbf{specific group} \quad Claim type: evaluative opinion \\
Stereotype: \textbf{present} \quad Vilification: \textbf{present} \\
Dehumanization: absent \quad Extreme Language: absent \\
Lack of Empathy: \textbf{present} \quad Invalidation: \textbf{present} \\
Decision basis: The text vilifies socialism by framing it as an invalid or illegitimate belief and invalidates the opposing viewpoint by dismissing it as ``illadvised''; the conditional acceptance structure (``I can deal with X, but not Y'') constitutes Lack of Empathy. \\
Final Answer: \textbf{1} \\
\hline
\end{tabular}
\caption{Qualitative comparison of SFT (Mistral-Nemo baseline) vs.\ best DPO model (299 DeepSeek-R1-filtered pairs, $\beta$=0.3) on a false-negative example from the test set. Bold fields differ between the two outputs. SFT misreads the implicit ideological framing as a personal anecdote and predicts 0. DPO recovers the correct label (1) by identifying Lack of Empathy and Invalidation, though it also flags Stereotype and Vilification as present, which are absent in the gold annotation.}
\label{tab:qual-example}
\end{table}

\subsection{Hyperparameters}
Post-submission experiments use Mistral-Nemo-Instruct-2407 as the base model. We attribute its stronger response to reasoning-based SFT — relative to Qwen2.5-7B-Instruct — to its larger parameter count (12B vs.\ 7B), explicit multilingual instruction-following training, and greater capacity to jointly learn a structured output schema and the classification objective. Larger models are known to benefit more from chain-of-thought style supervision \cite{wei2022chain}.

Table~\ref{tab:sft-hparams} reports the SFT hyperparameters.

\begin{table}[h!]
\small
\raggedright
\begin{tabular}{p{0.48\linewidth}p{0.44\linewidth}}
\hline
Parameter & Value \\
\hline
\textbf{Max Length} & 1024 \\
\textbf{Train Batch Size} & 1 \\
\textbf{Eval Batch Size} & 4 \\
\textbf{Gradient\newline Accumulation Steps} & 8 \\
\textbf{Learning Rate} & 5e-5 \\
\textbf{Num Epochs} & 3 \\
\textbf{Warmup Ratio} & 0.0 \\
\textbf{Precision} & BF16 \\
\textbf{Optimizer} & adamw\_torch \\
\textbf{Class Weights} & [1.0, 1.5] \\
\textbf{LoRA Rank (r)} & 8 \\
\textbf{LoRA Alpha} & 16 \\
\textbf{LoRA Dropout} & 0.05 \\
\textbf{LoRA Target Modules} & q\_proj, k\_proj, v\_proj, o\_proj \\
\hline
\end{tabular}
\caption{Pre-submission SFT hyperparameters (Qwen2.5-7B-Instruct). Class weights [1.0, 1.5] were used in both pre- and post-submission runs but did not massively improve performance over unweighted training (see Section~\ref{sec:results}).}
\label{tab:sft-hparams}
\end{table}

Table~\ref{tab:dpo-hparams} provides the complete DPO hyperparameter configuration.

\begin{table}[!htbp]
\small
\raggedright
\begin{tabular}{p{0.48\linewidth}p{0.44\linewidth}}
\hline
Parameter & Value \\
\hline
\textbf{Max Length} & 1024 \\
\textbf{Max Prompt Length} & 512 \\
\textbf{Train Batch Size} & 1 \\
\textbf{Gradient\newline Accumulation Steps} & 8 \\
\textbf{Learning Rate} & 5e-6 \\
\textbf{Num Epochs} & 2 \\
\textbf{$\beta$ (submitted)} & 0.1 \\
\textbf{$\beta$ (post-submission)} & 0.3 \\
\textbf{Warmup Ratio} & 0.0 \\
\textbf{Precision} & BF16 \\
\hline
\end{tabular}
\caption{DPO hyperparameters. Pre-submission uses Qwen2.5-7B-Instruct with $\beta$=0.1; post-submission uses Mistral-Nemo-Instruct-2407 with $\beta$=0.3.}
\label{tab:dpo-hparams}
\end{table}

\subsection{DPO Subset Sweep}
\label{sec:subset-sweep}

To examine whether the FP:FN ratio in the preference pair set affects DPO performance, we construct the subsets using stale preference pairs. We subsampled the class-0 (FP-correcting) pairs while keeping all class-1 (FN-correcting) pairs fixed, yielding three smaller pair sets. All experiments use the same hyperparameters (Mistral-Nemo, $\beta$ = 0.1, 2 epochs, lr = 5e-6) on the SFT Rejudged Sonnet checkpoint.

\begin{table}[h]
\centering
\small
\setlength{\tabcolsep}{4pt}
\begin{tabular}{lcccc}
\hline
Configuration & FP:FN & Total & mF1 & Recall \\
\hline
SFT baseline & --- & --- & 0.8097 & 0.827 \\
Original 190/70 & 73:27 & 260 & 0.8012 & 0.747 \\
Subset 150/70   & 68:32 & 220 & 0.8024 & 0.745 \\
Subset 120/70   & 63:37 & 190 & 0.8005 & 0.779 \\
Subset 100/70   & 59:41 & 170 & 0.7986 & 0.739 \\
\hline
\end{tabular}
\caption{DPO subset sweep on the stale unfiltered preference pairs: effect of subsampling class-0 (FP-correcting) pairs while holding class-1 (FN-correcting) pairs fixed at 70. Section~\ref{sec:post-submission}), confirming that pair quality and volume rather than FP:FN ratio alone drive the performance gap. Matching the test-set class distribution (120/70, 63:37) did not improve performance.}
\label{tab:subset-sweep}
\end{table}

 The 150/70 split achieved the best DPO result (0.8024), but sits much below the best-performing DPO system (0.8162). This gap persists despite the more balanced pair ratio, reinforcing the conclusion that LLM-judge quality filtering — not pair composition alone — is the key factor separating the best DPO configurations from the rest.

\subsection{Label-Only SFT: Training Details}
\label{sec:label-only-sft}

Label-only supervised fine-tuning trains the model to predict the 
final binary classification label directly, without generating 
intermediate reasoning steps. This serves as an important baseline 
for assessing the contribution of structured rationale generation.

\paragraph{Training Data.}
We use the full official English training split (3,222 examples) 
with the original annotations, without any re-validation or 
relabeling.

\paragraph{Input Format.}
Each training example follows a simplified two-field template:
\begin{verbatim}
Input:
<text>
Final Answer:
<label>
\end{verbatim}

\paragraph{Training Configuration.}
We fine-tune with LoRA using the same 
adapter configuration as the reasoning SFT baseline (rank 8, 
alpha 16, dropout 0.05, attention projections only). Standard 
causal language modelling loss is applied across all tokens in 
the sequence; no loss masking is used. Key hyperparameters are 
listed in Table~\ref{tab:label-only-hparams}.

\begin{table}[h]
\centering
\small
\begin{tabular}{ll}
\hline
Parameter & Value \\
\hline
\textbf{Base Model} & Qwen/Qwen2.5-7B-Instruct \\
\textbf{Training Data} & 3,222 samples (original) \\
\textbf{Max Length} & 1024 \\
\textbf{Batch Size} & 8 \\
\textbf{Gradient Accumulation} & 1 \\
\textbf{Learning Rate} & 5e-5 \\
\textbf{Num Epochs} & 10 \\
\textbf{Precision} & BF16 \\
\textbf{Optimizer} & AdamW \\
\textbf{LoRA Rank (r)} & 8 \\
\textbf{LoRA Alpha} & 16 \\
\textbf{LoRA Dropout} & 0.05 \\
\textbf{LoRA Target Modules} & q\_proj, k\_proj, v\_proj, o\_proj \\
\textbf{Loss} & Standard causal LM \\
\hline
\end{tabular}
\caption{Label-only SFT hyperparameters (Qwen2.5-7B-Instruct).}
\label{tab:label-only-hparams}
\end{table}

\subsection{SFT training data generation prompt}
\label{sec:sft-prompt}
The supervised fine-tuning dataset was generated using the Gemma 3 27B model served via Ollama's inference framework \cite{gemmateam2025gemma3technicalreport}. The prompt below is an excerpt from the template employed during data generation.
A very similar prompt was used to generate the DPO dataset with varying decoding temperatures; see Section~\ref{sec:dpo-preference-pairs}.

\begin{Verbatim}[breaklines=true]
Input:
<copy the input text verbatim>

Reasoning:
Target referenced: <specific 
individual / specific group / none>
Claim type: <factual description / moral judgment / evaluative opinion / call to action / other>
Manifestations present:
- Stereotype: <present / absent>
- Vilification: <present / absent>
- Dehumanization: <present / absent>
- Extreme Language and Absolutism: <present / absent>
- Lack of Empathy or Understanding: <present / absent>
- Invalidation: <present / absent>
Decision basis:
<one factual sentence explaining how the listed manifestations determine whether the text is polarized>

Final Answer: <output ONLY 0 or 1>
\end{Verbatim}

\subsection{Preference Pair Generation: Methods and Prompts}
\label{sec:pair-generation}

Across all experiments, preference pairs were constructed using the 
same mixed-outcome extraction procedure: for each training input, 
multiple completions were sampled at varied decoding temperatures 
to increase output diversity, classified as CORRECT, FP, or FN 
relative to the ground truth label, and paired according to the 
ranking CORRECT $\succ$ FP $\succ$ FN. The inference model used 
for completion generation differed across experimental stages.

\paragraph{Submitted System.}
Completions were sampled using Gemma~3 27B 
~\cite{gemmateam2025gemma3technicalreport} served via Ollama at 
varied temperatures. Preference pairs were then extracted from 
examples with mixed outcomes across completions.

\paragraph{Post-Ranking Qwen Experiments.}
The completions were sampled using GPT OSS 120B using 
the OpenRouter API  at 
temperatures 0.6, 0.9, and 1.2, generating six completions per 
example (two prompt strategies $\times$ three temperatures). 
Prompt~A argued in favor of a polarized label; Prompt~B argued 
against it.

Examples where all completions had the same outcome were excluded.

\paragraph{Post-Submission Mistral-Nemo Experiments.}
The completions were sampled with Llama-3.3-70B~\cite{grattafiori2024llama3herdmodels} from the Mistral-Nemo SFT checkpoint at 
various temperatures using the same two-prompt strategy. The 
resulting candidate pairs were then filtered using 
DeepSeek-R1~\cite{deepseekai2025deepseekr1} as 
an LLM judge, yielding the final 62:38 FP:FN balanced pair set 
used for DPO training at $\beta$=0.3.

\paragraph{Rejudged Sonnet Judging Prompt.}
Each training example was evaluated using the following prompt,
submitted to Claude~3.5 Sonnet. Reasoning for each example was first
generated by GPT OSS 120B Nitro; the judging prompt below was then
used to assess whether the generated reasoning supported the original
gold label.

\begin{Verbatim}[breaklines=True]
You are auditing generated rationale data for political
polarization training.
Task: Decide if this sample is acceptable for training data.
Acceptable means:
1. Reasoning is coherent and supports the final label.
2. Final Answer is 0 or 1 and matches GOLD_LABEL.
3. Required structure is present: Input, Reasoning,
   Final Answer.
4. Reasoning includes these slots in substance:
   * Target referenced
   * Claim type
   * Manifestations present (6 fields)
   * Decision basis
5. Claim type should be one of: factual description /
   moral judgment / evaluative opinion / call to action /
   other
Important:
* Do NOT fail only for minor formatting issues (e.g., one
  newline instead of two before Final Answer).
* Focus on semantic correctness and training usefulness.
Return strict JSON only with exactly keys:
\{"valid": bool, "confidence": 1-5, "errors": ["error_code"],
"notes": "short"\}
Allowed error codes: missing_section, missing_slot,
invalid_claim_type, invalid_label, label_mismatch,
contradictory_reasoning, weak_decision_basis,
format_violation

INPUT_TEXT: \{input_text\}
GOLD_LABEL: \{gold_label\}

GENERATED_SAMPLE: \{generated_sample\}
\end{Verbatim}

Examples were retained in the rejudged set only when Claude~3.5 Sonnet returned \texttt{"valid": true}.

\end{document}